\documentclass[sigconf]{acmart}

\AtBeginDocument{%
  \providecommand\BibTeX{{%
    \normalfont B\kern-0.5em{\scshape i\kern-0.25em b}\kern-0.8em\TeX}}}

\copyrightyear{2021}
\acmYear{2021}
\setcopyright{acmcopyright}\acmConference[MM '21]{Proceedings of the 29th ACM International Conference on Multimedia}{October 20--24, 2021}{Virtual Event, China}
\acmBooktitle{Proceedings of the 29th ACM International Conference on Multimedia (MM '21), October 20--24, 2021, Virtual Event, China}
\acmPrice{15.00}
\acmDOI{10.1145/3474085.3475337}
\acmISBN{978-1-4503-8651-7/21/10}
\acmSubmissionID{968}

\newcommand{\etal}{\textit{et al}.}
\newcommand{\ie}{\textit{i}.\textit{e}.}
\newcommand{\eg}{\textit{e}.\textit{g}.}

\usepackage{enumitem}
\usepackage{algorithmicx,algorithm}
\usepackage[noend]{algpseudocode}
\usepackage{multirow}
\usepackage{subcaption}
\usepackage[switch]{lineno}  %
\newcommand\blfootnote[1]{%
  \begingroup
  \renewcommand\thefootnote{}\footnote{#1}%
  \addtocounter{footnote}{-1}%
  \endgroup
}

\begin{document}
\fancyhead{} 

\title{Instance-wise or Class-wise?\\ A Tale of Neighbor Shapley for Concept-based Explanation}

\author{Jiahui Li$^1$, Kun Kuang$^{1*}$, Lin Li$^{1}$, Long Chen$^{2\dagger}$, Songyang Zhang$^{3}$, Jian Shao$^1$, Jun Xiao$^1$}

\affiliation{%
\institution{$^1$ Zhejiang University, $^2$Columbia University, $^3$University of Rochester}
\city{}
\country{}
}
\email{{jiahuil,kunkuang,mukti}@zju.edu.cn,zjuchenlong@gmail.com}
\email{szhang83@ur.rochester.edu,jshao@zju.edu.cn,junx@cs.zju.edu.cn}

\begin{abstract}
Interpreting model knowledge is an essential topic to improve human understanding of deep black-box models.
Traditional methods contribute to providing intuitive instance-wise explanations which allocating importance scores for low-level features (\eg, pixels for images).
To adapt to the human way of thinking, one strand of recent researches has shifted its spotlight to mining important concepts.
However, these concept-based interpretation methods focus on computing the contribution of each discovered concept on the class level and can not precisely give instance-wise explanations.
Besides, they consider each concept as an independent unit, and ignore the interactions among concepts. To this end, in this paper, we propose a novel COncept-based NEighbor Shapley approach (dubbed as CONE-SHAP) to evaluate the importance of each concept by considering its physical and semantic neighbors, and interpret model knowledge with both instance-wise and class-wise explanations. Thanks to this design, the interactions among concepts in the same image are fully considered. Meanwhile, the computational complexity of Shapley Value is reduced from exponential to polynomial. Moreover, for a more comprehensive evaluation, we further propose three criteria to quantify the rationality of the allocated contributions for the concepts, including coherency, complexity, and faithfulness.
Extensive experiments and ablations have demonstrated that our CONE-SHAP algorithm outperforms existing concept-based methods and simultaneously provides precise explanations for each instance and class.
\blfootnote{$^*$ Kun Kuang is the corresponding author.}
\blfootnote{$^\dagger$ This work started when Long Chen at Zhejiang University.}
\end{abstract}

\ccsdesc[500]{Computing methodologies~Machine learning}
\ccsdesc[500]{Computing methodologies~Supervised learning by classification}


\keywords{Model interpretation; Shapley value; Instance-wise explanation; Class-wise explanation}
\maketitle


\section{Introduction}
Deep neural networks have demonstrated remarkable performance in many data-driven and prediction-oriented applications~\cite{he2016deep,huang2017densely,zhuang2020next}, and sometimes even perform better than humans. 
However, their most significant drawback is the lack of interpretability, which makes them less attractive in many real-world applications. 
When relating to the moral problem or the environmental factors that are uncertain such as crime judgment~\cite{deeks2019judicial}, financial analysis~\cite{bracke2019machine}, and medical diagnosis~\cite{singh2020explainable}, it is essential to mine the evidence for the model's prediction (interpret model knowledge) to convince humans.
Thus, investigating how to interpret model knowledge is of paramount importance for both academic research and real applications.
\nocite{he2016deep}
\nocite{szegedy2016rethinking}
\nocite{huang2017densely}

\begin{figure}[t]
    \centering
    \includegraphics[width=0.85\linewidth]{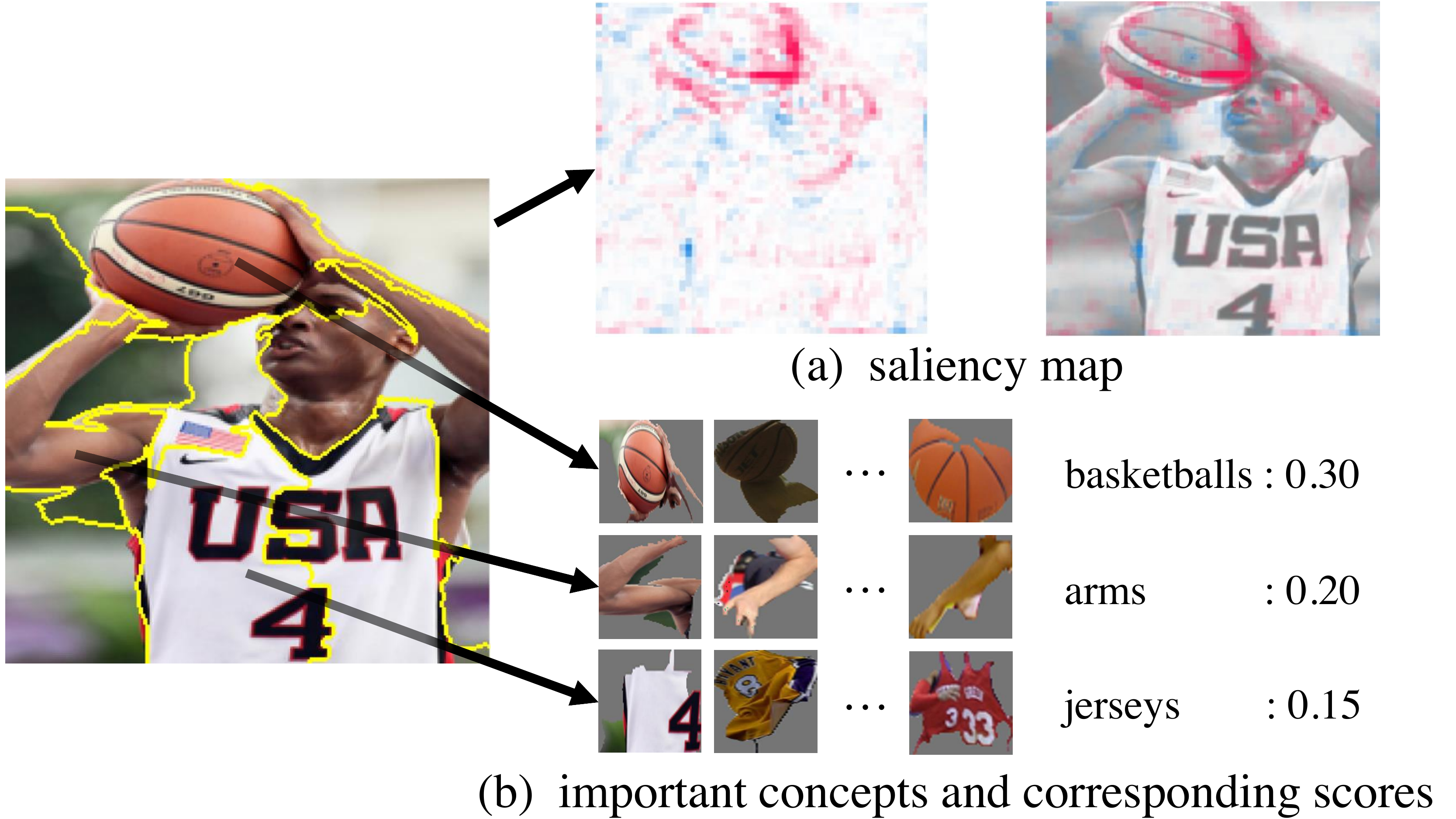}
    \caption{Interpreting the knowledge of a \emph{classification model} to recognize an image of \emph{basketball} by (a) feature-based explanation and (b) concept-based explanation.}
    \vspace{-3mm}
    \label{fig:introduction}
\end{figure}

The mainstream approaches to interpret model knowledge are \textbf{\emph{feature-based}} methods~\cite{ancona2019explaining,lundberg2017unified,ribeiro2016should,schwab2019cxplain,shrikumar2017learning,sundararajan2016gradients,wang2020scout,liang2019knowledge,binder2016layer,datta2016algorithmic,fashen2020interpret,kun2020causal,zhu2020dark,pan2020multiple}, which provide \textbf{instance-wise} explanation. They allocate the importance scores for each individual feature in each instance (\eg, each pixel in an image). 
Based on these importance scores, a saliency map that reflects the accordance for a model's decision intuitively for each instance is provided to improve humans' trusts. For example, in Figure~\ref{fig:introduction}(a), those red areas indicate the pixels which contribute most to the model to classify the image as basketball. 
However, those feature-based interpretations are not consistent with human understanding ~\cite{kim2018interpretability}, hence cannot help more on human decision and inference. 
Humans understand an image always based on high-level concepts, such as segments of \texttt{basketballs}, \texttt{arms} and \texttt{jerseys} as shown in Figure~\ref{fig:introduction}(b), rather than low-level pixels. 

To bridge the gap between human understanding and model interpreting, some \textbf{\emph{concept-based}} methods~\cite{ghorbani2019towards,kim2018interpretability,yeh2020completeness,wu2020towards,goyal2019explaining,zhou2018interpretable,higgins2016beta} which provide \textbf{class-wise} explanation have been proposed recently.
A concept can be a color, texture, or a group of similar segments that is easy for humans to understand.
These methods focus on mining a set of meaningful and representative concepts for an explained class and assigning importance scores according to their contributions to this class. Then, prototypes of the most important concepts found by the model are enumerated to convince humans.
However, all of the existing concept-based methods ignore differentiating the importance of concepts on each instance in a class. 
For example, \texttt{steering wheel} is a key concept for a model to recognize a car, these methods assigned the same importance score for \texttt{steering wheel} on all images, even for those without \texttt{steering wheel} visually. 
Hence, they lack the capacity for the instance-wise interpretation of model knowledge. Meanwhile, these methods regard each concept as an independent component, ignoring the interactions among concepts.
For example, \texttt{guitar's string} is an important concept, but without the participation of the \texttt{guitar's body}, the model can not distinguish whether the object in the image is a guitar or a violin.
Thus, it is not appropriate to calculate the contributions for each individual without considering its collaborators.

Hence, we are still facing the following challenges in interpreting model knowledge based on the concept to increase human understanding and trust in models:
(i) \textbf{Class-wise and instance-wise explanations}. Class-wise explanations interpret the decision boundary of the model, while instance-wise explanations show the unique importance of concepts for each instance. Both of them are important and necessary to increase human understanding and trust in models. 
(ii) \textbf{Interactions among concepts}. Concepts are far from independent, they could be physically interacted (each segment cooperates with its adjacent areas) or semantically interacted (each segment cooperates with its semantically similar areas).
(iii) \textbf{Evaluation of concept}. As stated in ACE~\cite{ghorbani2019towards}, a good concept should satisfy the properties of meaningfulness, coherency, and importance, but how to quantify those properties of discovered concepts is still a problem.

To address these challenges in interpreting model knowledge based on the concept for increasing human understanding and trust in models, we first propose a novel COncept-based NEighbor Shapley (dubbed as CONE-SHAP) method to approximate the importance of each concept with considering its interactions with the physical and semantic neighbors.
Meanwhile, we interpret the model knowledge from both instance-wise and class-wise, \ie, we calculate the neighbor Shapley Value of each possible concept in each instance and the top-ranked concepts give an explanation on the instance level (instance-wise) and average the importance of concepts over all instances in a class and selecting top-ranked concepts for explaining on the class level (class-wise).
Finally, to comprehensively quantify the discovered concepts and their importance score for interpreting model knowledge, we propose three criteria: coherency, complexity, and faithfulness. We validate our CONE-SHAP with extensive experiments. The results demonstrate that our algorithm outperforms both feature-based and concepts-based methods in interpreting model knowledge.

The main contribution we made in this paper can be summarized as follows:
\begin{itemize}
\item We investigate the problem of how to interpret model knowledge with concept-based explanations from both instance-wise and class-wise.
\item We propose a novel CONE-SHAP method to approximate the Shapley Value of each segment with considering its physical and semantic neighbors
\item We propose three criteria (coherency, complexity, and faithfulness) to comprehensively quantify the quality of the discovered concepts and their importance scores.
\item Extensive experiments show our CONE-SHAP outperforms the existing feature-based and concept-based methods on both class-wise and instance-wise explanations on models.
\end{itemize}

\section{Related Work}
\subsection{Feature-based Explanation}
Feature-based explanation methods focus on assigning importance scores for the features in an instance (\eg, pixels for an image, words for a text). 
These methods can be further categorized into several branches: (i) Perturbation-based methods~\cite{fong2017interpretable,zeiler2014visualizing,zintgraf2017visualizing,ancona2019explaining}, they quantify each feature by measuring the variation of outputs when masking or disturbing that feature while keeping the remaining fixed; (ii) Backpropagation-based methods~\cite{lundberg2017unified,bach2015pixel,sundararajan2016gradients}, 
they compute the importance scores of all features through a few times gradient-related operations; (iii) Model-based methods~\cite{schwab2019cxplain,ribeiro2016should}, they employ an explainable model to interpret the original model locally or train an extra deep network, which can output the feature importance directly.
These methods can visualize the important features in each instance for explanation (\eg, highlight the pixels in an image). But these feature-based explanations are always not consistent with human understanding~\cite{kim2018interpretability}. Moreover, these methods above assume all the features are independent and underestimate the interactions among features.

\subsection{Concept-based Explanation}
Humans always understand an object based on high-level concepts rather than fine-grained features. Concepts are defined as a group of similar prototypes which can be understood easily by humans.
Given a human-defined concept, Been~\etal~\cite{kim2018interpretability} proposes TCAV to quantify the significance of the concepts in a class different from other categories based on trained linear classifiers. 
ACE~\cite{ghorbani2019towards} employs super-pixel segmentation and cluster methods for mining concepts automatically, and then adopted TCAV~\cite{kim2018interpretability} for the concept-based explanation.
ConceptSHAP~\cite{yeh2020completeness} defines the notion of completeness score to measure the semantic expression ability of a concept and utilized Shapley Value~\cite{shapley1953value} to find a complete set of concepts. 
All of these concept-based methods explain deep models on the class-level, but underestimate the local structure of each instance.

\subsection{Shapley Value for Explaining Models}
Shapley Value~\cite{shapley1953value,shapley1988shapley,kumar2020problems} originated from cooperative game theory and is the best way to distribute benefits fairly by considering the contributions of various agents. Thus, some of the recent studies borrow the idea from Shaley Value to interpret deep neural network~\cite{ghorbani2019data,ancona2019explaining,chen2018shapley}.
Nevertheless, its computational complexity increases exponentially with the number of participating members. Since there may exit a large number of features in an explained instance, it is exorbitant for a computer to calculate the Shapley Value.
To overcome this challenge, the approximation of Shapley Value~\cite{castro2009polynomial,owen1972multilinear,fatima2008linear,jones2010multilinear,li2021shapley}, is used as a substitution which might slightly break some properties of Shapley Value.
Ghorbani~\etal~\cite{ghorbani2020neuron} utilize a sample-based method to estimate the Shapley Value for each neuron in the
model.
Ancona~\etal~\cite{ancona2019explaining} also adopt a sample-based method to approximate the Shapley Value and allocate the importance for all features via one-time forward propagation.
Chen~\etal~\cite{chen2018shapley} propose L-Shapley and C-Shapley to approximate Shapley value in a graph structure. 
Though these approximations have made excellent performance in many tasks, most of them focus on features and play roles on the instance level.

Our method overcomes the above shortcomings and explains the model knowledge from both class-level and instance-level. 
The details will be demonstrated in  \emph{Section~\ref{sec:method}}.

\section{Preliminaries of Shapley Value} \label{sec:preliminaries}
Shapley Value~\cite{shapley1953value,chen2018shapley,goyal2019explaining} is designed in cooperative game theory to distribute gains fairly by considering the contribution of several players working in a big coalition.
Assume a game consists of $N$ players and they cooperate with each other to achieve a common goal.
let $u(\cdot)$ represents the utility function to measure the contributions made by an arbitrary set of players. For a particular player $i$, let $S$ be an arbitrary set that contains player $i$, and $S \backslash \{i\}$ represents the set with the absence of $i$, then the marginal contribution of $i$ in $S$ is defined as:
\begin{equation}
m(i,S)=u(S)-u(S \backslash \{i\}).
\end{equation} 

\noindent The Shapley Value of player $i$ is defined as:
\begin{equation}
\phi_u(i)=\frac{1}{N}\textstyle{\sum}^N_{k=1} \frac{1}{\binom{N-1}{k-1}}\textstyle{\sum}_{S \in S_k(i)}{m(i,S)},
\end{equation}
where $S_k(i)$ denotes a set with size $k$ that contain the player $i$.
Shapley Value is the unique value to satisfy the following properties:
\begin{itemize}[leftmargin=*]
    \item \textbf{Efficiency:} The value of the whole union $u(\{1,...,N\})-u(\emptyset)$ is equal to the sum of the Shapley Values of all of the players $\sum{^{N}_{i=1}}\phi_u(i)$.
    
    \item \textbf{Symmetry:} If $m(i,S)=m(j,S)$ for all subsets $S$ then $\phi_u(i)=\phi_u(j)$.
    
    \item \textbf{Dummy:} If $m(i,S)=0$ for all subsets $S$ then $\phi_u(i)=0$.

    \item \textbf{Additivity:} Let $u$ and $w$ represent the associated utility functions, then $\phi_{u+w}(i)=\phi_u(i)+\phi_w(i)$ for every $i \in N$.
    
    \item \textbf{Coherency:} Given another value function $m'(i,\cdot)$ to measure the marginal contribution of $i$, if $m(i,S)\geq m'(i,S)$ for all subsets $S$, then $\phi_u(i)\geq \phi^{'}_u(i)$.

\end{itemize}
\begin{figure*}[ht]
    \centering
    \includegraphics[width=0.85\linewidth]{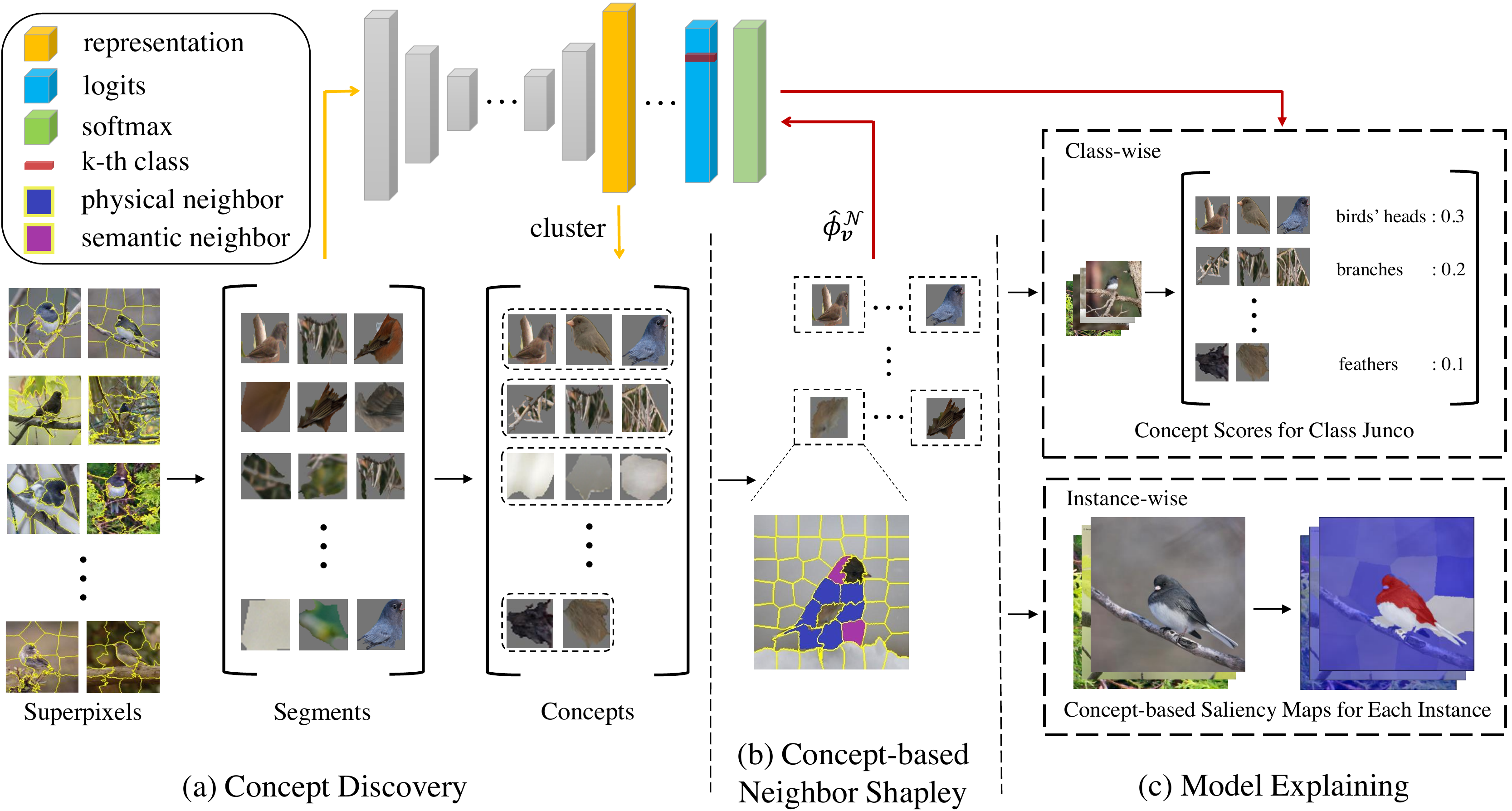}

    \caption{The framework of our CONE-SHAP. \textbf{(a)} Segmenting all the samples via super-pixel operation and then cluster the segments into concepts according to their representations in the explained model. \textbf{(b)} Calculating the importance scores for the concepts in each instance and allocate the concepts' scores for an explained class by averaging the importance of concepts over all instances in this class. \textbf{(c)} Providing instance-wise explanation via concept-based saliency maps and provide class-wise explanation via concepts with importance scores.}
  \label{fig:framework}
\end{figure*}

\section{Evaluating Concepts}
\subsection{Problem Formulation} 
Given a trained neural network
$f$, and a set of input examples
${X^{k}}=\{x^{k}_{1}, x^{k}_{2}, ...,x^{k}_{n}\}$
to be explained for a target class 
$k$, where
$x^{k}_{i}$
denote the $i^{th}$ example of class
$k$, concept-based methods aim at finding out a set of meaningful concepts
${C^{k}}=\{C^{k}_{1}, C^{k}_{2}, ...,C^{k}_{m}\}$
as well as importance scores
${SC^{k}}= \{SC^{k}_{1}, SC^{k}_{2}, ...,SC^{k}_{m}\}$
for each concept according to its contribution to the model for 
class $k$. For convenience, we omit the superscript of the above symbols and rewrite 
${X^{k}}$
as ${X}=\{{x_{1},x_{2}, ...,x_{n}}\}$,
${C^{k}}$
as
$C=\{C_{1},C_{2}, ...,C_{m}\}
$ and ${SC^{k}}$
as
${SC}=\{SC_{1},SC_{2}, ...,SC_{m}\}$  without causing ambiguity.
\subsection{Criteria for Evaluating Concept Score}
\noindent\textbf{Previous Methods and Shortcomings.}
ACE~\cite{ghorbani2019towards} utilizes \emph{smallest sufficient concepts} (SSC) and \emph{smallest destroying concepts} (SDC) to quantify the quality of the concepts.
SSC means to find a set of concepts which are enough for the model to make a prediction, and it is used to measure the representation ability of the extracted concepts. SDC means to look for a set of concepts that will cause a poor prediction when these concepts are removed, and it reflects the necessity of the concepts for a model's decision. 
ConceptSHAP~\cite{yeh2020completeness} proposes \emph{completeness} to measure the expression ability of concepts.
These existing metrics for evaluating concepts only consider part of the concepts' properties for a model's decision and lack the capacity to measure the concepts from different aspects.
Thus, in this paper, we quantify three criteria (coherency, complexity, and faithfulness) to comprehensively evaluate concepts.

\noindent\textbf{High Coherency.}
A concept with a higher score should have a stronger ability to express the semantic of its original inputs, we define this property as \emph{high coherency}. Let
$f$ be a pre-trained model, which maps a input
$x$ to an output
$f(x)$.
And let
$h(\cdot)$
be the function that maps input 
$x$
to a representation layer. We define the similarity between the $i^{th}$ concept and its original input as:
\begin{equation} \label{equation:coherency}
\eta(C_i;X) = \frac{1}{|C_i|} {\sum}^{|C_i|}_{j=1}{\text{Sim}(h(x|c_{i,j}), h(c_{i,j}))},
\end{equation}
where
${\left | C_i \right |}$ denotes the numbers of segments in the inputs which belongs to concept $i$,
${c_{i,j}}$ denotes the $j^{th}$ segment of the $i^{th}$ concept, 
${x}{\vert}c_{i,j}$ denotes the original input instance which contains
${c_{i,j}}$ and ${\text{Sim}}(\cdot,\cdot)$ denotes the function which measure the similarity between two tensor such as cosine similarity.
Assuming that 
$\{1:k\}$ represent the first $k$
value of a variable, thus
${C}\{1:k\}$ was the top-$k$ concepts with the highest contribution, and 
${SC}\{1:k\}$ are their corresponding importance scores, the top-$k$ coherency is defined as:
\begin{equation}
\zeta_k(SC,\eta;C,X)=\text{corr}(SC\{1:k\},\eta(C;x)\{1:k\}),
\end{equation}
where 
$\text{corr}(\cdot,\cdot)$ 
represents correlation coefficient between two variables. 

\noindent\textbf{Low Complexity.}
We want the distribution of the scores of different concepts should be distinguished from each other as much as possible. 
The concepts' scores are considered complex if they are the same or very close to each other. First, we normalize the scores of top-$k$ concept, and the normalized concept score of the $i^{th}$ concept is defined as:
\begin{equation}\Tilde{SC_i}=\frac{SC_i}{\sum^{k}_{j=1}SC_j}.
\end{equation}
Note that
$\Tilde{SC_i}$ can also be treated as valid probability distribution, then we define the top-$k$ complexity of the concepts as the entropy of
$\hat{S_i}$:
\begin{equation}
    \xi_k(\Tilde{SC})=-{\sum}^{k}_{i=1}{\Tilde{SC_i}}\ln{\Tilde{SC_i}}.
\end{equation}

\noindent\textbf{High Faithfulness.}
The change of the model's outputs when the concepts are removed or set to baseline should be correlated with the concepts' scores, we define this property as \emph{high faithfulness}. Let 
$X\backslash\{C_i\}=\{x_1\backslash\{C_i\},x_2\backslash\{C_i\},...x_n\backslash\{C_i\}\}$ denotes the input samples with the absence of all the segments belongs to the $i^{th}$ concept, where
$x_j\backslash\{C_i\}$ denotes the $j^{th}$ processed input instance by setting all the segments in the $i^{th}$ concept as zero or baseline. Let:
\begin{equation}
  \varphi(C_i;X)=\frac{1}{n}{\sum}^{n}_{j=1}(f(x_j)-f(x_j\backslash C_i)),
\end{equation}
where $\varphi(C_i;X)$ represents the degree of the degradation of the model's performance. Since the number of the segments of the concepts various, we normalize
$\varphi(C_i;X)$
as:
\begin{equation}
\Tilde{\varphi}(C_i;X)=\frac{1}{|C_i|}\varphi(C_i;X).
\end{equation}
Then the faithfulness of the top-$k$ concepts' scores is defined as:
\begin{flalign} \label{equation:faithfulness}
&\theta_k(SC,\varphi;C,X)=\text{corr}(SC\{1:k\},\Tilde{\varphi}(C;X)\{1:k\}).
\end{flalign}

\section{Explaining Model via CONE-SHAP} \label{sec:method}
To bridge the gap between model decision and human understanding, we propose a post-hoc approach, namely COncept-based NEighbor Shapley (CONE-SHAP), to interpret the decision procedures of any deep neural network from both instance-wise and class-wise levels with human-friendly concepts. The framework of our CONE-SHAP is exhibited in Figure~\ref{fig:framework}.
In order to interpret model knowledge for a target class, our method first extracts concepts automatically and then computes an importance score for each concept according to our proposed CONCE-SHAP. Finally, the concept-based saliency maps of each instance are given for instance-wise explanation, and the concepts' contributions for the whole class are given for class-wise explanation.

\subsection{Concept Discovery}
Concepts are defined as prototypes that are understandable for humans~\cite{wu2020towards,yeh2020completeness}. 
In computer vision tasks, it can be a color, texture, or a group of similar segments, and in natural languages, it can be a group of sub-sentences with the same meaning.
Since there are no user-defined concepts in real-world applications, a method to discover concepts automatically is needed. 
Following ACE~\cite{ghorbani2019towards}, we assume a set of semantically similar images' segments as a visual concept. 
To collect such kind of concept, firstly, a super-pixel method is performed on each sample of the inputs $X$, thus we get a dozen segments from a particular class. 
Then, these segments are clustered into $m$ different concepts according to their representations computed by model 
$f$. We denote the concepts as $C=\{C_1, C_2,...,C_m\}$, 
where
$C_i=\{c_{i,1},c_{i,2},...,c_{i,|C_i|}\}$, 
$c_{i,j}$ denotes the $j^{th}$ segment in the $i^{th}$ concept,
$\vert C_i \vert$ is the number of the segments belong to the $i^{th}$ concept, and $m$ denotes the number of the discovered concepts.

Notice that such an approach relies on a good super-pixel method because the objects of different sizes occupy different proportions in the image which may keep some meaningful concepts from being discovered. For example, the `jersey' in Fig~\ref{fig:introduction} is divided into two parts, though they are a whole.
To avoid the meaningful concepts are missed, the multi-grained super-pixel method with three different sizes (small, medium, large) is adopted thus we get the segments with multi-resolution.
The details and the hyperparameters will be discussed in \emph{Section~\ref{sec:exp}}. 
\subsection{Concept-based Neighbor Shapley}
To measure the contribution of a segment in an instance for an explained model, we apply a counterfactual method which considers how the prediction of the model will change if this segment is absent.
For classification tasks, let
$g$
be the last layer before the \emph{softmax} operation and 
$g_k$ represents the logit values of class $k$.
The value of a sgement
$s$ of class $k$ for the model is calculated as:
\begin{equation}
 v_k(s)=g_k(x)-g_k(x\backslash\{s\}).
\end{equation} For convenience, we denote
$
v_k(s)
$
as 
$
v(s)
$.

\noindent\textbf{Shapley Value.}
We consider all the $N$ segments in an image as a union, and each of them is a player. Given a particular player $i$, let 
$S$ be a subset that contains player $i$ and 
$S \backslash \{i\}$ denotes the subset without the participation of $i$, then
the contribution of $i$ to the subset $S$ is computed as:
\begin{equation} \label{eq:conterfactual}
\Delta v(i,S)=v(S)-v(S \backslash \{i\}).
\end{equation} 

\noindent When we treat $v(\cdot)$ as the utility function, then $\Delta v(\cdot)$ becomes the marginal contribution of the Shapley value.
Thus, the Shapley Value of player $i$ is the weighted average of marginal contribution in all of the subsets:
\begin{equation}\label{eq:sv}
\phi_v(i)=\frac{1}{N}{\sum}^N_{j=1} \frac{1}{\binom{N-1}{j-1}}{\sum}_{S \in S_j(i)}{\Delta v(i,S)},
\end{equation}
where
$S_j(i)$ denotes the set with size $j$ that contains the $i^{th}$ segment.

However, the computational complexity for Shapley Value increases exponentially as the number of players rises~\cite{kumar2020problems,sundararajan2019many,frye2019asymmetric}. Since each image contains more than a hundred segments, it is costly for a computer to calculate the truly Shapley Value.
Thus, recent studies replaced the truly Shapley Value with its approximation~\cite{chen2018shapley,ancona2019explaining,ghorbani2020neuron,covert2020understanding} in different situations.

\noindent\textbf{Approximation of Shapley Value.}
Inspired by~\cite{frye2019asymmetric}, the players can be treated as the nodes of a fully connected graph, where any two players are connected since they might have correlations during a game.
In the application of image classification, the segments of an image can also be treated as nodes, but each node only connects with its neighbors.
Here, we define the neighbors $\mathcal{N}(i)$ of the $i^{th}$ segment are those segments which are adjacent to it (physical neighbors) or belong to the same concept as it (semantic neighbors).
Based on the assumption that participants which are not the neighbors of 
$i$ hardly affect its contribution for a model's inference procedure, the Shapley Value of $i$ in \emph{Equation}~\ref{eq:sv} can be approximated as:
\begin{equation} \label{eq:nsv}
     \phi^\mathcal{N}_v(i)=\frac{1}{|\mathcal{N}(i)|}{\sum}^{|\mathcal{N}(i)|}_{j=1} \frac{1}{\binom{|\mathcal{N}(i)|-1}{\quad j-1}}{\sum}_{^{ i \in S}_{S \subseteq \mathcal{N}(i)}}{\Delta v(i,S)}.
\end{equation}

Considering that a segment may contain a large amount of neighbors in an instance, we adopt sample-based method to estimate $\phi^\mathcal{N}_v(i)$ in order to further reduce the computational costs.
Concretely, we first sample $k$ nodes from $\mathcal{N}(i)$ and denote it as
$\mathcal{N}_k(i)$, and then compute the Shapley Value in the $\mathcal{N}_k(i)$. This procedure will repeat $M$ times, and we take the average of these results as the COncept-based NEighbor Shapley Value (CONE-SHAP) of $i$:
\begin{equation} \label{eq:asv}
   \hat{\phi}^\mathcal{N}_v(i)=  \frac{1}{M|\mathcal{N}_k(i)|}{\sum}^M_{t=1}{\sum}^{|\mathcal{N}_k(i)|}_{j=1} \frac{1}{\binom{|\mathcal{N}_k(i)|-1}{\quad j-1}}{\sum}_{^{i \in S}_{S \subseteq \mathcal{N}_k(i)}}{\Delta v(i,S)}.
\end{equation}

Next, we will introduce how to employ the approximation of Shapley Value from \emph{Equation}~\ref{eq:asv} to interpret model knowledge from both instance-wise and class-wise.
And more discussion between CONE-SHAP and other approximations of Shapley Value can be found in Appendix.


\begin{figure*}[h]

    \centering
    \includegraphics[width=0.80\linewidth]{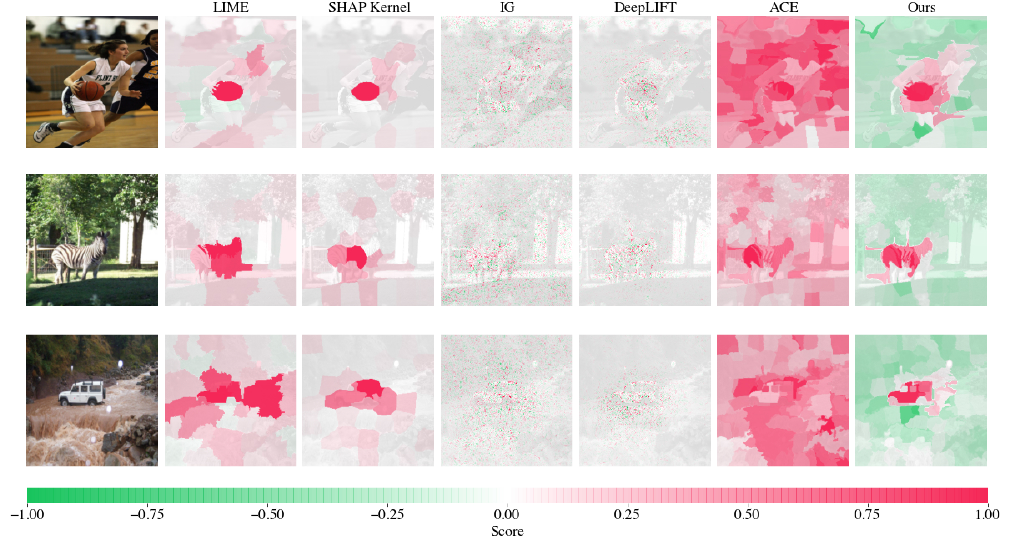}

    \caption{The comparison of the saliency maps of our method and existing popular methods (feature-based and concept-based). Three images in ImageNet are selected as examples and pre-trained densenet-121 is used as the explained model. }
    \label{fig:compare_instance}

\end{figure*}


\begin{figure}[h]    
\includegraphics[width=0.82\linewidth]{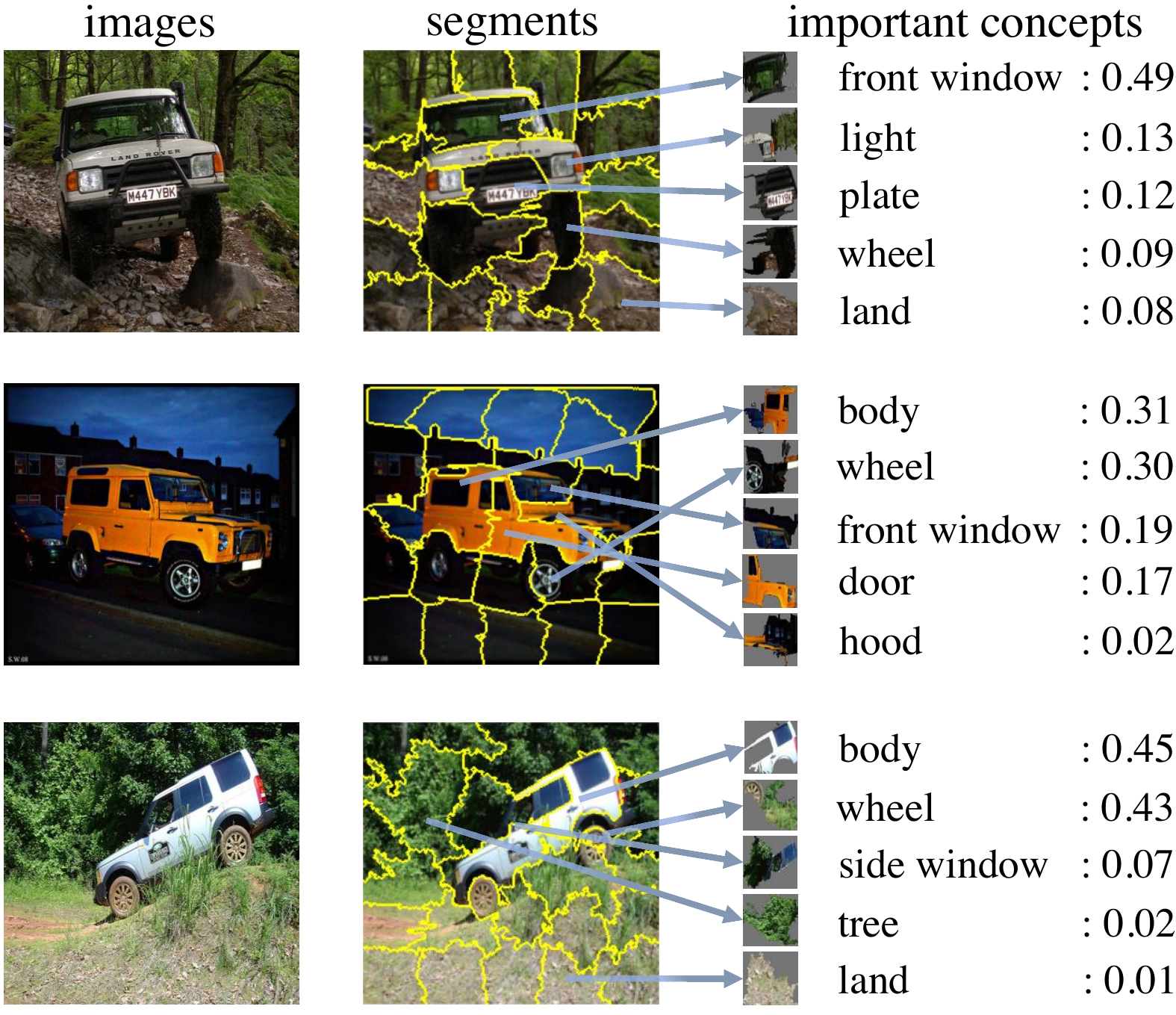}
\vspace{-2mm}
\caption{Show cases of class \texttt{jeep} in ImageNet and their 5 most important explained by our method. The same concept might have different importance across instances (images).}
\vspace{-6mm}
\label{fig:instance_concepts}
\end{figure}

\subsection{Model Explaining}
\noindent\textbf{Instance-wise Explanation.}
In order to help users understand the basis for a model's reasoning procedure intuitively, we provide concept-based saliency maps to interpret model knowledge on the instance-level.
The contribution of each segment of each instance is assigned according to its CONE-SHAP Value 
$
\hat{\phi}^\mathcal{N}_v(i)
$. Compared to perturbation-based methods which explain a model in fine-grained features, our CONE-SHAP focuses on the concept-based explanation, which is more human-friendly.

\noindent\textbf{Class-wise Explanation.}
To interpret model knowledge on the class level, our method distributes the concept scores to indicate which concept contributes more to the model's prediction on the explained class.
A concept is considered important if all of its belongings own a high Shapley Value.
Since we have gotten a group of possible concepts in the concept discovery procedure, for a concept
$C_i$, we compute its score by averaging all of the approximate Shapley Values of its segments:
\begin{equation}
    SC_i=\frac{1}{|C_i|}{\sum}_{c_{i,j} \in C_i}\hat{\phi}^\mathcal{N}_v(c_{i,j} ),
\end{equation}
where $c_{i,j}$ represents the segment belongs to concept $C_i$.
\subsection{Analysis on CONE-SHAP}
\noindent\textbf{Analysis of Computational Complexity.}
The Shapley Value of a segment is computed according to \emph{Equation}~\ref{eq:sv}. 
When we interpret an target instance, let $n$ be the average number of segments per instance, then the computational complexity of \emph{Equation}~\ref{eq:sv} is $O(2^n)$. 
When we approximate \emph{Equation}~\ref{eq:sv} via \emph{Equation}~\ref{eq:nsv} which assume that the participated players only cooperate with its neighbors, the computational complexity downgrade to $O(2^{|\mathcal{N}|})$. $\mathcal{N}$ is the average number of the neighbors for a segment which is much more smaller than $n$.
When $\mathcal{N}$ is large, our proposed CONE-SHAP estimates \emph{Equation}~\ref{eq:nsv} via \emph{Equation}~\ref{eq:asv}, then the computational complexity becomes $O(M2^k)$.
Since both $k$ and $M$ are small constants, the final computational complexity can be written as $O(d)$, where $d$ is a positive integer.

\noindent\textbf{Analysis of the Approximation.}

\newtheorem{thm}{\bf Theorem}[]
\begin{thm}\label{thm:error bound}
When the mutual information between a player $i$ and any of its neighbors is bounded by $\epsilon$, the expected error between Equation~\ref{eq:nsv} and Equation~\ref{eq:sv} is bounded by $2\epsilon$:
\begin{equation}
\mathbb{E}|\phi^\mathcal{N}_v(i) - \phi_v(i)|\leq 2\epsilon.
\end{equation}
\end{thm} 

\begin{thm}\label{thm:converge}
Equation~\ref{eq:asv} converges to Equation~\ref{eq:nsv} as $M \to \infty$.
\end{thm} 
\noindent We leave the proofs in Appendix.

\section{Experiments} \label{sec:exp}
In this section, we first discuss the experimental settings and the hyperparameters we adopt.
Then, we demonstrate the interpretation ability of our method from both instance-wise and class-wise. 
Meanwhile, we evaluate the efficiency of the top-ranking concepts. 
Finally, the rationality of the allocated concepts' scores is measured according to the criteria that we proposed. 

\begin{figure*}[h]
    \begin{subfigure}{.63\textwidth}
    \includegraphics[width=0.5\linewidth]{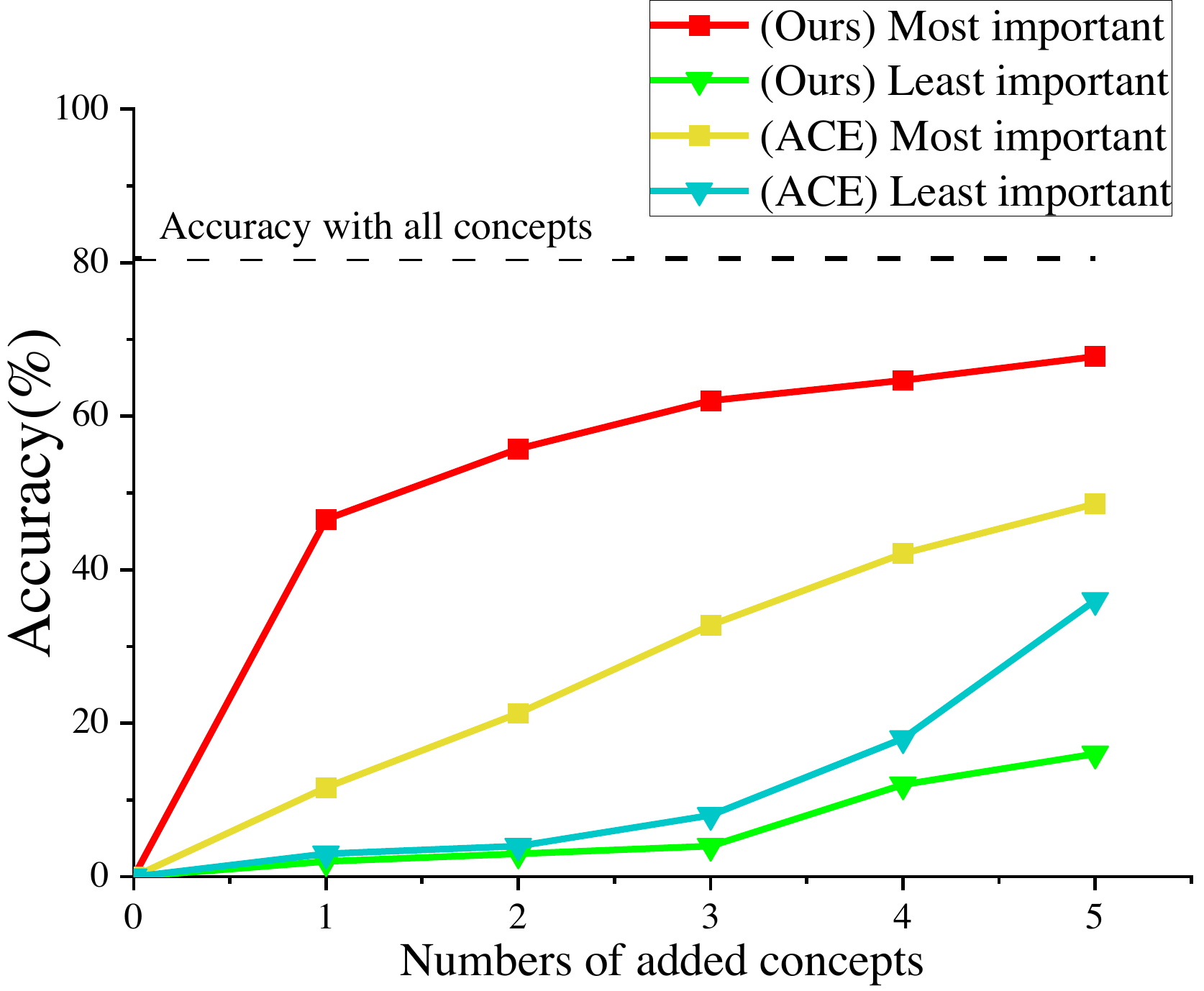}
    \hspace{2mm}
    \includegraphics[width=0.5\linewidth]{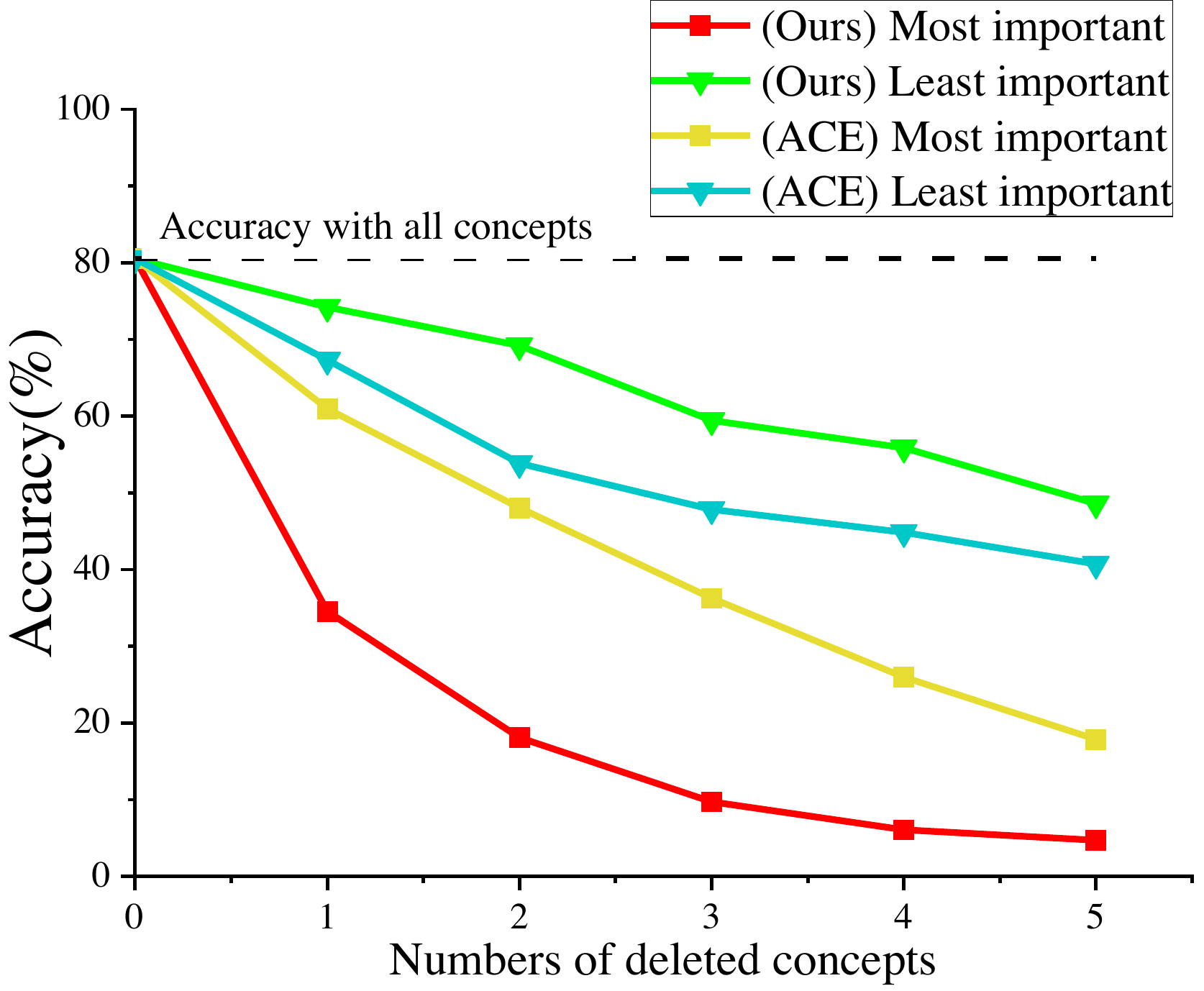}
    \caption{}
     \vspace{-3mm}
    \end{subfigure}
    \hspace{+5mm}
    \begin{subfigure}{.26\textwidth}
    \includegraphics[width=\linewidth]{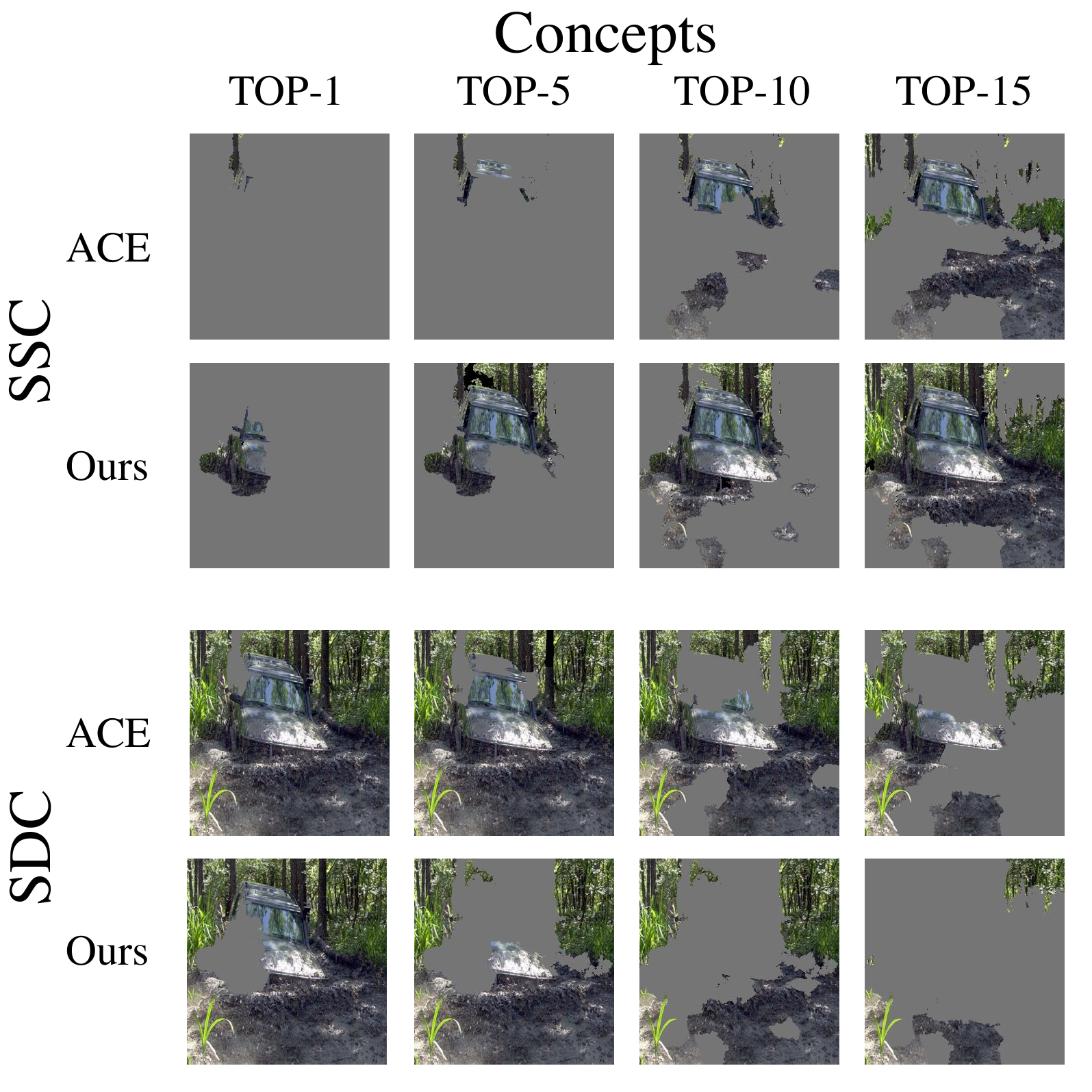}
    \caption{}
     \vspace{-3mm}
    \end{subfigure}
    \caption{The comparison of performance of the top-ranked concepts computed by our CONE-SHAP and ACE. (a) The changes of the model's performance when the most important 5 concepts are added/removed gradually. (b) An example of removing/adding the top-ranked concepts of an image in class \texttt{jeep}.}
    \vspace{-2mm}
    \label{fig:SSC}
\end{figure*}

\subsection{Experimental Settings}
Our method can be applied to any task without further training. In order to demonstrate the superiority of our method intuitively, we focus on image classification in this paper. 

\noindent\textbf{Models and Dataset.}
We interpret two most commonly used official models Densenet-121~\cite{huang2017densely} and Inception-V3~\cite{szegedy2016rethinking} which pretrained on the ImageNet dataset~\cite{deng2009imagenet} in the experiments. The former is used for instance-wise explanation and the latter is used for class-wise explanation.

\noindent\textbf{Settings for Concept Discovery.}
For each explained instance, we first picked out all of the samples that belong to the same class, and then mine out the candidate concepts for this class by segmenting these instances and using K-means method to cluster the fragments. 
Since the meaningful concepts for an object are usually no more than 20, we set the cluster center to 20 for each class.
As we mentioned before, the discovery of the concepts relies on a good segmentation method, in order to avoid neglecting meaningful segments, we perform multi-resolution super-pixel for each image. Concretely, SLIC~\cite{achanta2012slic} is used to segment input samples for speed, and each instance is divided into three granularities with the resolutions of 15, 50, 80 segments separately. Thus, the concepts with different sizes are captured.

\noindent\textbf{Settings for Feature Extraction.}
As for the features we used to cluster the image fragments, we resize the segments to the size of the original images via bicubic interpolation, and feed them to the explained model to get the representations of the middle layer of the neural network. We adopt \emph{ave-pool} layer in Inception-V3~\cite{szegedy2016rethinking} and \emph{global ave-pool} layer in Densenet-121~\cite{huang2017densely} as the representation layer respectively.

\noindent\textbf{Settings for Approximated Shapley Value.}
For the importance of each segment, we compute the approximated Shapley Value according to \emph{Equation}~\ref{eq:asv}. To make the balance between the computational complexity and the interpretation performance, we set the hyperparameter $k$ as 5 and $M$ as 1 to sample while calculating the approximated Shapley Value. The choice of these hyperparameter will be discussed in  \emph{Subsection~\ref{sec:5-3}}. 


\subsection{Instance-wise Explanation}

For instance-wise explanation, we compare our method to LIME~\cite{ribeiro2016should}, SHAP kernel~\cite{lundberg2017unified}, Integrated Gradients (IG)~\cite{sundararajan2016gradients}, DeepLIFT~\cite{shrikumar2017learning}, and ACE~\cite{ghorbani2019towards} via  saliency maps. Specially, ACE is a concept-based method, and can't provide saliency maps directly, thus we got the saliency maps by assigning the TCAV~\cite{kim2018interpretability} scores to its segments. 
We demonstrate ACE here in order to show that the other concept-based methods lack the capability to explain the model from instance-wise.

\noindent\textbf{Explanation with Saliency Maps.}
We provide coarse-grained saliency maps with the same size of the inputs to indicate which part of an instance is more important.
Since we have got the approximated Shapley Value of each segment of each image with three different granularities, we treat these values as the scores in saliency maps, and then we get three saliency maps focus on different scales. For each instance, we average the multi-grained saliency maps to get the final saliency map.
Our method provides concept-based saliency maps which are coarse-grained but more human-friendly. 

Figure~\ref{fig:compare_instance} shows the comparison of the saliency maps between our CONE-SHAP and baselines, where the importance scores of each unit are scaled between -1 
and 1 by dividing the absolute value of the maximum number, and the units with scores below zero were set to green and beyond zero were set to red.
There exist more green areas in our method because we consider the correlation between each area adequately, and the backgrounds and the meaningless segments hardly cooperate with the others.
Compared with methods that distribute importance scores for each pixel, our concept-based saliency maps are easier for humans to understand.
ACE~\cite{ghorbani2019towards} focus on interpreting concepts from class-wise thus fails to provide precise explanations for each instance. 

\noindent\textbf{Different Importance of Concepts on Different Instances.}
Intuitively, even the same concept might have different importance for different instances/images. Based on the value of CONE-SHAP on each segment, our method can estimate the importance of each concept on each instance as follows: Firstly, we find out all of the concepts in an instance. Then, for each concept, we compute the CONE-SHAP value of each segment that belongs to the concept with \emph{Equation}~\ref{eq:asv}. Finally, we can estimate the importance of the concept by sum up the CONE-SHAP values of its segments.

Figure~\ref{fig:instance_concepts} exemplifies the interpretation of our CONE-SHAP for a model to recognize \emph{jeep}. In the class-level, our method shows the concept of \texttt{bodies}, \texttt{windows}, \texttt{plates}, and \texttt{wheels gears} are very important for the class of jeep, while in a specific instance/image, different concepts have different importance. For example, in the first picture of Figure~\ref{fig:instance_concepts}, the most important concept is \texttt{front window} with the score $0.49$, but we can not see the \texttt{font window} in the third picture, hence the most important concept in the third image changes to \texttt{jeep's body}. And \texttt{wheel} is important in all of the three pictures but it plays different roles in different instances.  

\begin{table*}[]
\caption{\textbf{(a)} The changes of the prediction accuracy when gradually removing the top-$k$ important concepts (the lower the better). \textbf{(b)} The changes of the prediction accuracy when gradually adding the top-$k$ important concepts (the higher the better). }
\vspace{-1mm}
\subfloat[]{\setlength\tabcolsep{7pt}  

\centering
\begin{tabular}{l|c c c c c}
\hline
\multirow{2}{*}{Methods}    & \multicolumn{5}{c}{SDC (removing top-$k$ important concepts)} \\
& top-1     & top-2     & top-3     & top-4     & top-5 \\
\hline
ACE         & 60.97     & 48.07     & 36.27     & 25.98     & 17.88 \\
Ocllusion   & 34.58     & 19.93     & 14.10     & 11.61     & 8.38  \\
SHAP (MC)    & 34.48     & 18.58     & 13.15     & 10.42     & 5.91  \\ 
\hline
\textbf{Ours} & \textbf{34.15}   & 18.11 & \textbf{9.75} & \textbf{6.08} & \textbf{4.73} \\
Ours ($w/o$ PN) & 36.56 & \textbf{17.72} & 12.42    & 9.44      & 4.92  \\
Ours ($w/o$ SN) & 34.75     & 19.45     & 12.29     & 6.08      & 4.86  \\
\hline

\end{tabular}}
\hspace{4.6mm}
\subfloat[]{\setlength\tabcolsep{7pt}  

\centering\begin{tabular}{l|c c c c c}
\hline
\multirow{2}{*}{Methods}    & \multicolumn{5}{c}{SSC (adding top-$k$ important concepts)} \\
& top-1     & top-2     & top-3     & top-4     & top-5     \\ 
\hline
ACE         & 11.63     & 21.28     & 32.78     & 42.09     & 48.57     \\ 
Ocllusion   & 44.40     & 51.35     & 55.96     & 59.67     & 61.36     \\ 
SHAP (MC)    & 44.56     & 52.36     & 55.43     & 60.36     & 63.36     \\ 
\hline
\textbf{Ours} & \textbf{46.56} & \textbf{55.76} & \textbf{62.01} & 64.68 & \textbf{67.78} \\
Ours ($w/o$ PN) & 44.02     & 54.86     & 59.62     & 63.36     & 66.00     \\
Ours ($w/o$ SN) & 45.56     & 54.59     & 60.50 & \textbf{64.90} & 66.41    \\ 
\hline
\end{tabular}}

\label{tab:SSC SDC}
\end{table*}

\begin{table*}[]
\caption{\textbf{(a)} The mean changes of the quality of the concepts when the number of sample times $M$ increases when we classify 5 classes selected randomly in the test set of ImageNet. \textbf{(b)} The mean changes of the quality of the concepts when the number of sampled neighbors $k$ increases when we classify 5 classes selected randomly in the test set of ImageNet. }
\label{tab:sample analysis}
\vspace{-1mm}
\subfloat[]{\setlength\tabcolsep{7pt}  

\centering
\begin{tabular}{l|c c c}
\hline
Metrics    & M=1       & M=2       & M=3    \\ \hline
SSC-most   & 43.20     & 44.00     & 42.00  \\ \hline
SSC-least  & 6.40      & 6.80      & 5.20   \\ \hline
SDC-most   & 33.20     & 34.00     & 33.20  \\ \hline
SDC-least  & 82.80     & 85.20     & 86.00  \\ 
\hline
\end{tabular}}
\hspace{13.6mm}
\subfloat[]{\setlength\tabcolsep{7pt}  
\centering\begin{tabular}{l|c c c c c}
\hline
Metrics    & k=1       & k=2       & k=3       & k=4       & k=5   \\ \hline
SSC-most   & 31.60     & 34.00     & 38.80     & 40.80     & 42.80 \\ \hline
SSC-least  & 11.20     & 8.80      & 8.00      & 8.00      & 5.60  \\ \hline
SDC-most   & 41.20     & 40.40     & 36.80     & 36.40     & 32.40 \\ \hline
SDC-least  & 83.20     & 83.80     & 82.40     & 84.00     & 84.80 \\ 
\hline
\end{tabular}}

\end{table*}



\subsection{Class-wise Explanation} \label{sec:5-3}

The compared baselines \footnote{We did not compare with TCAV~\cite{kim2018interpretability} and ConceptSHAP~\cite{yeh2020completeness}, since they rely on the dataset with human-labeled concepts.} for class-wise explanation include ACE~\cite{ghorbani2019towards}, SHAP(MC), which approximates the Shapley Value with Monte Carlo sampling, and Ocllusion which compute the importance for each segment according to \emph{Equation}~\ref{eq:conterfactual}.

\noindent\textbf{Validating the Performance of Concepts.} To measure the top-$k$ important concepts on the explained model, we employ the same metrics (\ie, SSC and SDC) as ACE~\cite{ghorbani2019towards}, where SSC/SDC represents the accuracy of model prediction when we add/remove the most important concepts on the image.
We use official Inception-V3~\cite{szegedy2016rethinking} without any further training as the explained model. 
20 classes in ImageNet are selected randomly to explain, and we calculate SSC and SDC for each class separately and take an average. 

Table~\ref{tab:SSC SDC} reports the changes of the prediction accuracy by gradually removing/adding the top-$k$ important concepts, where the method \emph{$w/o$ PN} and \emph{$w/o$ SN} refer to the ablations from our method by removing the Physical Neighbor (PN) and Semantic Neighbor (SN) in \emph{Equation}~\ref{eq:nsv} \& \ref{eq:asv}, respectively. From the results, we can conclude that (i) By removing/adding the top-$k$ important concepts, our method makes the model achieve the lowest/highest accuracy. This is because our method can estimate the importance of each concepts more precisely than baselines. (ii) Both of the ablations (\emph{$w/o$ PN} and \emph{$w/o$ SN}) would lead a worse performance on our method, which indicates that both of the physical and semantic neighbor considered in our CONE-SHAP method is necessary. We also plot Figure~\ref{fig:SSC} to clearly demonstrate the advantages of our CONE-SHAP compared with ACE.
From the Figure~\ref{fig:SSC}(a), we have following observations: (i) By adding/deleting the most important concepts from the image, our method changes (improve/reduce) the model accuracy more remarkably than baseline ACE. (ii) By adding/deleting the least important concepts, our method have less influence on the model accuracy than baseline. Moreover, we show an example of adding/removing the most important concepts of an image in Figure~\ref{fig:SSC}(b).

\noindent\textbf{Analysis of the Hyperparameters for Approximating Shapley Value.}
We approximated the truly Shapley Value via sampling from the neighbors as depicted in \emph{Equation}~\ref{eq:asv}.
A large $M$ and $k$ will bring pressure to computation costs, and a small $M$ and $k$ will lead to inaccurate estimates. 
We performed extensive experiments to find moderate $M$ and $k$.
For choosing $M$, we select five classes randomly to interpret. We fix $k$ to 5 and increase $M$ gradually to compute \emph{SSC-most}, \emph{SSC-least}, \emph{SDC-most}, and \emph{SDC-least} of the selected classes. 
The results shown in Table~\ref{tab:sample analysis}(a) demonstrate that in our settings, set $M$ to 1 is enough to approximate the Shapley Value.
Then for choosing $k$, we set $M$ to 1 and increase $k$ gradually. We find that 5 neighbors is sufficient to reach relatively high performance. 
Table~\ref{tab:sample analysis}(b) shows the mean results of 5 classes selected randomly to exhibit how the performance changes with the growth of $k$.
Thus, we set $M$ to 1 and $k$ to 5 in our experiments.

\begin{table}[t]
\caption{The quality of the allocated scores for the top-5 concepts on the criteria we proposed. }
\centering
\begin{tabular}{|l|ccc|}
\hline
\multirow{2}{*}{Methods}  & \multicolumn{3}{c|}{Criteria top-5} \\
& Coherency     & Complexity      & Faithfulness    \\ 
\hline
ACE              & 0.4090         & 1.5943          & 0.4516          \\ 
Ocllusion       & 0.7437          & 1.4051          & 0.9245          \\ 
SHAP (MC)   & 0.7602          & 1.3930          & 0.9325          \\
\hline
\textbf{Ours} & \textbf{0.9299}     & 1.2622          & \textbf{0.9542} \\
Ours ($w\backslash o$ PN)  & 0.8423          & 1.3135 & 0.9427          \\ Ours ($w\backslash o$ SN)    & 0.9235          & \textbf{1.2422}          & 0.9462          \\
\hline
\end{tabular}
\vspace{-7mm}
\label{tab:CCF table}
\end{table}

\noindent\textbf{Evaluation of Concept Scores.}
SSC and SDC merely take into account whether concepts are ranked rationally, and it is also necessary to evaluate the quality of the distributed scores.
Hence, we evaluated the scores computed by various methods based on the three criteria we proposed. We adopt Pearson correlation coefficient to compute the coherency in \emph{Equation}~\ref{equation:coherency} and faithfulness in \emph{Equation}~\ref{equation:faithfulness} , and the results are depicted in Table~\ref{tab:CCF table}. As the table shows, our CONE-SHAP obtains the best score on \emph{coherency} and \emph{faithfulness} that reach 0.93 and 0.95 respectively. That means the concept score and expressive ability are highly consistent, and the scores are highly correlated with its contributions to model performance. 
Besides, the low complexity reflects our scoring mechanism can distinguish concepts from each other very well.

\section{Conclusion}
In this paper, we investigate the problem of post-hoc explanation for deep neural networks in a human-friendly way.
First, we propose a method named CONE-SHAP which can explain any model from both instance-wise and class-wise without further training.
Especially, by considering the interaction neighbors, our CONE-SHAP downgrades the computational complexity of Shapley Value from exponential to polynomial.
Since there are no unified metrics to measure the performance of concepts' scores, we proposed three criteria to evaluate the scoring mechanism of concept-based explanation methods.
Extensive experiments demonstrate the superior performance of our method. 
Applying our method to real-world applications, users can easily understand the important shreds of evidence for a model's prediction so as to make a decision confidently.

\begin{acks}
This work was supported by the National Key Research and Development Project of China (No.2018AAA0101900), the National Natural Science Foundation of China (No. 61625107, U19B2043, 61976185, No. 62006207), Zhejiang Natural Science Foundation (LR19F020002), Key R \& D Projects of the Ministry of Science and Technology (No. 2020YFC0832500), Zhejiang Innovation Foundation(2019R52002), the Fundamental Research Funds for the Central Universities and Zhejiang Province Natural Science Foundation (No. LQ21F020020).
\end{acks}
\bibliographystyle{ACM-Reference-Format}
\bibliography{sample-base}

\end{document}